\documentclass[letterpaper]{article} 
\usepackage{aaai24}  
\usepackage{multicol}
\usepackage{multirow}
\usepackage{color}
\usepackage{amssymb}
\usepackage{bbding}
\usepackage{times}  
\usepackage{helvet}  
\usepackage{courier}  
\usepackage[hyphens]{url}  
\usepackage{graphicx} 
\usepackage{natbib}  
\usepackage{caption} 
\usepackage{algorithm}
\usepackage{algorithmic}
\usepackage{amsmath}
\usepackage{cite}
\usepackage{booktabs}

\usepackage{newfloat}
\usepackage{listings}
\DeclareCaptionStyle{ruled}{labelfont=normalfont,labelsep=colon,strut=off} 
\floatstyle{ruled}
\newfloat{listing}{tb}{lst}{}
\floatname{listing}{Listing}

\usepackage{bibentry}

\begin{document}
\nocopyright
\title{Dynamic V2X Autonomous Perception from Road-to-Vehicle Vision}
\author{
    Jiayao Tan\textsuperscript{\rm 1}\equalcontrib,
    Fan Lyu\textsuperscript{\rm 2}\equalcontrib,
    Linyan Li\textsuperscript{\rm 3},
    Fuyuan Hu\textsuperscript{\rm 1}\thanks{Corresponding author.},
    Tingliang Feng\textsuperscript{\rm 4},
    Fenglei Xu\textsuperscript{\rm 1},
    Rui Yao
}
\affiliations{
    \textsuperscript{\rm 1}Suzhou University of Science and Technology,
    \textsuperscript{\rm 2}CRIPAC, MAIS, CASIA,\\
    \textsuperscript{\rm 3}Suzhou Institute of Trade Commerce,
    \textsuperscript{\rm 4}Tianjin University\\
    \{jiayaotan@post,\textsuperscript{\rm †}fuyuanhu@mail,xufl@mail\}.usts.edu.cn, fan.lyu@cripac.ia.ac.cn, lilinyan@szjm.edu.cn


%
}
\maketitle
\begin{abstract}

Vehicle-to-everything (V2X) perception is an innovative technology that enhances vehicle perception accuracy, thereby elevating the security and reliability of autonomous systems.
However, existing V2X perception methods focus on static scenes from mainly vehicle-based vision, which is constrained by sensor capabilities and communication loads.
To adapt V2X perception models to dynamic scenes, we propose to build V2X perception from road-to-vehicle vision and present \emph{Adaptive Road-to-Vehicle Perception} (AR2VP) method. 
In AR2VP, we leverage roadside units to offer stable, wide-range sensing capabilities and serve as communication hubs.
AR2VP is devised to tackle both intra-scene and inter-scene changes. 
For the former, we construct a dynamic perception representing module, which efficiently integrates vehicle perceptions, enabling vehicles to capture a more comprehensive range of dynamic factors within the scene.
Moreover, we introduce a road-to-vehicle perception compensating module, aimed at preserving the maximized roadside unit perception information in the presence of intra-scene changes. 
For inter-scene changes, we implement an experience replay mechanism leveraging the roadside unit's storage capacity to retain a subset of historical scene data, maintaining model robustness in response to inter-scene shifts.
We conduct perception experiment on 3D object detection and segmentation, and the results show that AR2VP excels in both performance-bandwidth trade-offs and adaptability within dynamic environments. Our code is available at: https://github.com/tjy1423317192/AP2VP
\end{abstract}


\subsection{Introduction}
The Vehicle-to-everything (V2X) technique \citep{63,8,9}, facilitating collaboration between vehicles and various other entities \cite{13,14}, is introduced as a popular means to enhance the perception system for intelligent driving \cite{2,AAAI4：dri}. 
\emph{However, existing V2X research predominantly centers around static data (barely no entity or scene change), which inadequately addresses the safety prerequisites of vehicles in dynamic traffic environments.}
To elaborate, a dynamic traffic environment encompasses two key facets:
(1) {Intra-scene variations}: This involve the variations within a scene, containing factors like pedestrians in motion and vehicles executing turns. 
(2) {Inter-scene variations}: This pertains to substantial changes in architecture and lane configurations between different scenes. 
An autonomous system should possess the capability to facilitate driving in dynamic environments, where inaccurate perception may potentially lead to traffic incidents.

\begin{figure}[t]
  \centering
  \includegraphics[width=\linewidth]{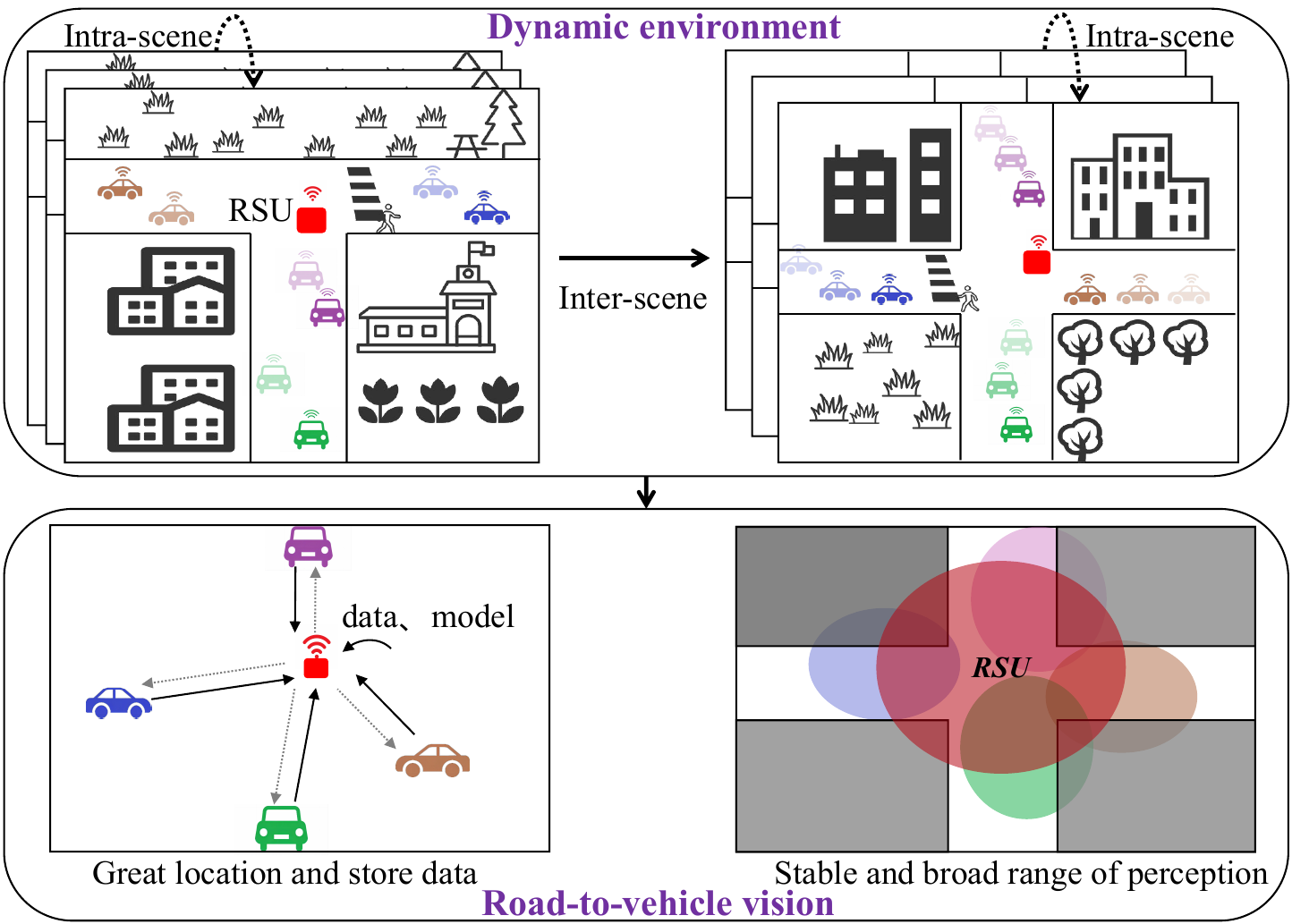}
  \caption{Dynamic Scene Advantage of RSU. RSU demonstrates stability in perception and possess geographical advantages in inter-scene changes, as opposed to the continual mobility of vehicles in intra-scene dynamics. With storage and communication capabilities, RSU stores old scenes data and model, enabling adaptation to inter-scene changes.}
  \label{Fig.1}
  \vspace{-15px}
\end{figure}

The primary limitation of existing V2X studies in accommodating dynamic environments stems from their heavy dependence on \textit{vehicle-based vision}, constrained by sensor capabilities and communication loads.
In this paper, we improve V2X perception from road-to-vehicle vision, which utilizes a stationary Road Side Unit (RSU) \cite{RSU1,RSU2} positioned at a fixed location, providing a more stable and expansive sensory coverage while minimizing redundant communication. 
As shown in Fig.~\ref{Fig.1}, within dynamic traffic scenarios, RSU serves as perception nodes situated at the heart of the traffic scene. 
This facilitates the collection of comprehensive perception data over a wider range and ensures more consistent perception as compared to the inherently mobile vehicles.
Furthermore, RSU boasts increased storage and computing capabilities, enabling rapid adaptation to evolving environments. 
\emph{To the best of our knowledge, our work represents a pioneering effort in harnessing the road-to-vehicle vision paradigm to enhance V2X perceptions within dynamic scenes.}

Motivated by this, this paper proposes \textbf{Adaptive Road-to-Vehicle Perception (AR2VP)} approach, which constructs a road-to-vehicle cooperative perception model tailored for dynamic environments.
First, to effectively handle intra-scene changes, we design a dynamic perception representing module and road-to-vehicle perception compensating module.
These modules collaboratively construct an adaptable graph catering to intra-scene changes, thereby enhancing vehicles' overall adaptability within dynamic scenarios.
Second, to effectively adapt to inter-scene changes, we present RSU experience replay, utilizing RSU storage capacity and integrating experience replay \cite{AAAI1：conti,60,61} techniques in continual learning \cite{58,58.1,58.2,58.3,59}, enabling vehicles to adapt to large-scale scene transitions beyond intra-scene changes. 
Our approach is validated on the tasks of scene segmentation and 3D object detection.
The results on V2X-Sim dataset~\cite{63} show good perception performance and the adaptability of AR2VP to dynamic scenes. 

Our contributions are three-fold:

\begin{itemize}
\item To the best of our knowledge, we are the first to investigate the adaptability of V2X to dynamic scenes. Accordingly, we also proposes AR2VP approach for dynamic V2X perception.
\item To address intra-scene changes, we design Dynamic Perception Representing module and Road-to-Vehicle Perception Compensating module. These modules tap into the perceptual insights from the road side, thereby bolstering the overall adaptability of vehicles within dynamic scenes.
\item To effectively handle intra-scene changes, we put forward the concept of RSU Experience Replay. This mechanism empowers vehicles to seamlessly adapt to substantial scene transitions that extend beyond the scope of mere intra-scene changes.

\end{itemize}

\section{Related Work}

\noindent
\textbf{Perception in V2X.}
V2X technology encompasses various forms of cooperative communication, including Vehicle-to-Vehicle (V2V) \cite{43(v2v)} and Vehicle-to-Infrastructure (V2I) \cite{45(v2i)}. 
For \textit{V2V} technology, Who2com \cite{19} exploits a handshake communication mechanism to determine which two vehicles should communicate for image segmentation. 
When2com \cite{18} introduces an asymmetric attention mechanism to decide when to communicate and how to create communication groups for image segmentation. V2VNet \cite{20} proposes multiple rounds of message passing on a spatial-aware graph neural network for joint perception and prediction in autonomous driving. 
DiscoNet \cite{60} proposes distilled collaboration graph with matrix-valued edge weights for adaptive perception, offering superior performance-bandwidth trade-off.
\citet{65} proposes a pose error regression module to learn to correct pose errors when the pose information from other vehicles is noisy.
For \textit{V2I} technology, the collaboration is between infrastructure and vehicles, which expands the vehicle's perception field. 
\emph{However, most of both existing V2V and V2I methods build V2X perception model from vehicle vision, which is insufficient in dynamic traffic environment.}
In this paper, we aim to construct road-to-vehicle vision to address the challenge of inadequate adaptability of collaborative perception models in dynamic environments.

\noindent
\textbf{Continual Learning.}
Continual learning is a commonly used approach for adapting to changing scenarios. It allows the model to continuously update itself while receiving new data, thereby accommodating various environmental changes. Some common methods in this context include regularization \cite{incremental,incremental1}, experience replay, and parameter freezing \cite{freez,freez1}. 
Traffic scenes are characterized by significant variability, where continual learning holds promise for application in complex and dynamic traffic environments. 
\emph{However, despite its successes in other domains, there is currently no research considering the utilization of continual learning for modeling V2X perception systems.} 
Traditional V2X technologies did not account for scene changes, resulting in the perception model experiencing forgetting phenomena \cite{ACL1：forget,ACL2：forget} when vehicles transition between different scenes. 
This paper investigates the potential of applying continual learning to model V2X perception systems, with the aim of better adapting to inter-scene changes, 
thereby enhancing the safety and reliability of the automomous.
\begin{figure*}
  \centering
  \includegraphics[width=\linewidth]{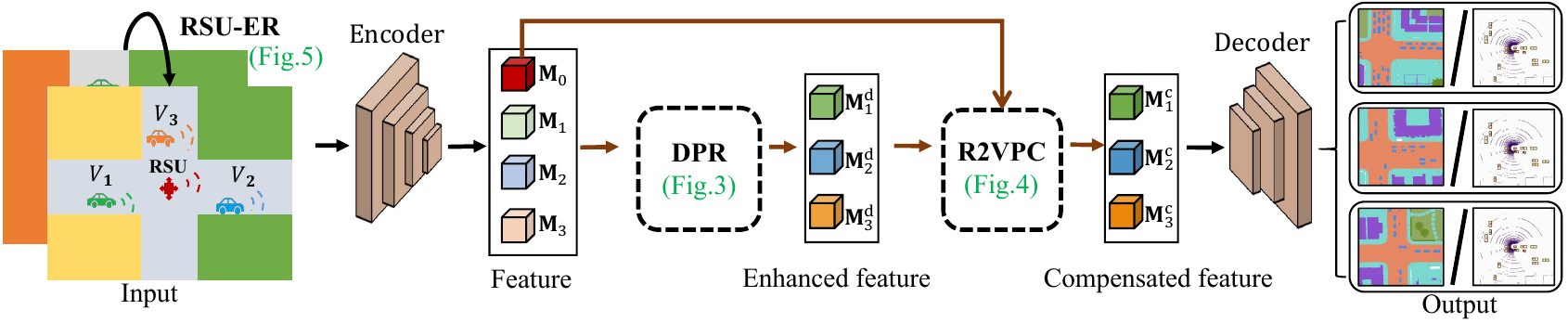}
\vspace{-15px}
  \caption{Overall perception model framework. For the vehicle (green), its BEV map ${\mathbf{V}_1}$ is encoded by the shared Encoder to obtain the feature map ${\mathbf{M}_1}$. Based on the Dynamic Perception Representing module, the neural information from other vehicles and RSU is aggregated to obtain the feature map ${\mathbf{M}_1^{\rm d}}$. It is compensated with the neural information ${\mathbf{M}_0}$ from the RSU to obtain the feature map ${\mathbf{M}_1^{\rm c}}$. The shared header outputs the result after the shared Decoder.}
  \label{Fig.2}
    \vspace{-15px}
\end{figure*}




\section{Adaptive Road-to-Vehicle Perception}
\subsection{Overview}

We study the V2X perception task with RSU placed on dynamic scenes.
In one scene, the V2X perception consists of an RSU and vehicles. 
The RSU and vehicles collect point cloud data, these input single-view point cloud can be converted to bird’s-eye-view (BEV) \cite{60} maps ${\mathcal V}={ \{\mathbf{V}_0,\mathbf{V}_1,\mathbf{V}_2,...,\mathbf{V}_i\}}$, where ${\mathbf{V}_0}$ for RSU and ${\mathbf{V}_i}$ for the $i$-th vehicle ($i>0$).
Existing V2X perception technique primarily focus on static data, which falls short of meeting the safety requirements in dynamic traffic environments: 
Intra-scene changes, such as pedestrians in motion and moving vehicles, along with inter-scene changes like transitions between extensive structures and road layouts across different locations, introduce disruptions to V2X perception, potentially compromising vehicle safety.

Motivated by this, this work considers to build a collaborative perception model from road-to-vehicle vision for sensing complex and dynamic traffic scenarios.
We name the method {Adaptive Road-to-Vehicle Perception} (AR2VP), as shown in Fig.~\ref{Fig.2}, where vehicles and RSU communicate and cooperate through a broadcast communication channel.
AR2VP considers to address two kinds of scene changes:
\textbf{Intra-scene changes:} we first design \emph{Dynamic Perception Representing} (DPR) module, utilizing RSU geographical and perceptual advantages to effectively integrate the perception from vehicles, enabling vehicles to capture a more comprehensive range of dynamic factors within the scene. Then, to further enhance vehicles perception capabilities in dynamic environments, we draw inspiration from residual \cite{AAAI2：res,AAAI3：res} techniques and propose \emph{Road-to-Vehicle Perception Compensating} (R2VPC) module. Leveraging RSU perceptual advantages, this module compensates the post-collaborative perception of vehicles, filling in intra-scene dynamic factors that overlooked by the vehicles, thereby further enhancing the overall adaptability of vehicles to dynamic environments. Lastly, to enable vehicles to adapt to large-scale scene transitions beyond intra-scene changes.
\textbf{Inter-scene changes:} we introduce \emph{RSU Experience Replay}. This combines RSU storage capability with experience replay techniques from continual learning, enabling AR2VP to adapt to inter-scene changes, ensuring reliable vehicles perception.

\subsection{Overcoming intra-scene changes}

\begin{figure}[t]
	\centering
	\includegraphics[width=\linewidth]{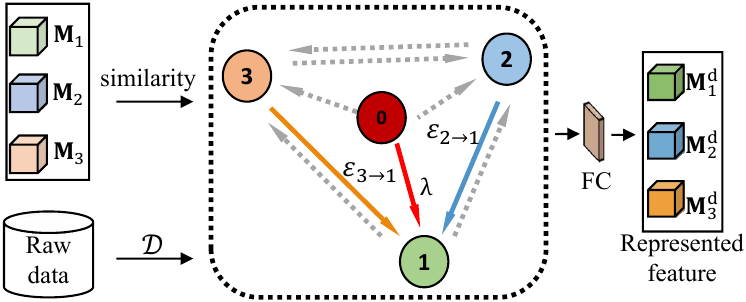}
    \caption{Dynamic Perception Representing. For the feature map $\mathbf{M}_1$, we calculate the weights $\mathcal{\xi}$ by combining feature similarity with the RSU-Vehicle distances $\mathcal{D} = \{d_1,d_2,...,d_i\}$ to construct a directed collaborative graph $\{\mathcal{M},\mathcal{\xi}\}$, followed by perceptual representation of the feature map $\mathbf{M}_i^{\rm d}$.}
    \label{Fig.3}
    \vspace{-15px}
\end{figure}

\noindent
\textbf{Dynamic Perception Representing.}
Vehicle perception varies with the changes of dynamic entities within the intra-scene. 
To effectively coordinating these dynamic factors, this paper propose a Dynamic Perception Representing (DPR) module, which constructs a directed collaborative graph $\mathcal{G} =\{\mathcal{M},\mathcal{\xi}\}$ that leverages the advantages of RSU to adapt to intra-scene changes (See Fig.~\ref{Fig.3}), where $\mathcal{M}={\Phi _{\rm shared}}(\mathcal{V})$ is encoded by the shared encoder ${\Phi _{\rm shared}(\cdot)}$ to generate the feature maps and $\mathcal{V}$ represents BEV maps. 
The collaborative graph has three stages for dynamic perception representation: 

\noindent
\emph{(1) Stage S1: position information transforming}. 
In this stage, each vehicle transfers the position to the RSU for interaction. RSU and each vehicle has its own independent position ${\mathcal P} = \{(x_0,y_0),(x_1,y_1),(x_2,y_2),...,(x_n,y_n)\}$. When selecting the position of the ${i}$-th vehicle for collaboration, we need to transform the position information of RSU $(x_0,y_0)$ into $(x_{0 \to i},y_{0 \to i})$ corresponding to the ${i}$-th vehicles using the position matrix:
\begin{equation} 
x_{0 \to i}={\mathbf{R}_i}{\mathbf{R}_0^\top}(x_0-x_i),\quad y_{0 \to i} ={\mathbf{R}_i}{\mathbf{R}_0^\top}(y_0-y_i), 
\end{equation}
where ${\mathbf{R}_i}$ denotes the rotation matrix of the ${i}$-th vehicle, which represents the orientation of coordinate system relative to the reference coordinate system. 
Note that ${\mathbf{R}_0}$ is RSU's rotation matrix. 
Then, we pass the converted position information to the S2 stage to obtain the edge weights.

\noindent
\emph{(2) Stage S2: Position-guided feature fusing}.
In this stage, each vehicle receives effective perception information from both RSU and other vehicles in the same scene. 
RSU, due to its unique geographical position, offers vehicles dynamic environmental adaptability. 
Therefore, based on the position information from Stage S1, we combine the relative distance between RSU and vehicles with the feature information between vehicles, and carry out effective collaborative perception. In the directed collaborative graph $\mathcal{G}$, to determine edge weights $\mathcal{\xi}$, we first obtain the distances $\mathcal{D} = \{d_1,d_2,...,d_i\}$ between vehicles and the RSU from the Stage S1:
\begin{equation} 
{d_i} = \sqrt{{(x_{0 \to i}-x_i)}^2+{(y_{0 \to i}-y_i)}^2}.
\end{equation}
Then, we associate the features of different vehicles. 
In other words, the matrix value of the edge weight from the ${2}$-th vehicle to the ${1}$-th vehicle $\xi_{2 \to 1}$:
\begin{equation} 
\xi_{2 \to 1} = \frac{d_2\cdot{\rm cos}({\mathbf{M}_1},{\mathbf{M}_2})}{\sum\nolimits_{i=2}^{N} {d_i\cdot{\rm cos}({\mathbf{M}_1},{\mathbf{M}_i})}},
\end{equation}
where ${\rm norm(\cdot)}$ represents set normalization, and ${\rm cos(\cdot)}$ represents feature similarity.
For the edge weights between RSU and vehicles, we use fixed weight $\lambda  = \frac{1}{N}$ to retain the stable perception information of RSU. Through Stage 2, we complete the construction of graph $\mathcal{G}$.

\noindent
\emph{(3) Stage S3: feature information aggregating}: 
In this stage, we utilize the directed collaborative graph constructed in Stage S2 to synergistically enhance the representation of each vehicle. The perception information of each vehicle and RSU is integrated to better capture dynamic entities, achieving a comprehensive perception of the entire environment. Specifically, each vehicle aggregates the normalized edge-weighted features of all other vehicles. The updated feature map of the ${i}$-th vehicles is ${\hat{\mathbf{M}}_i}$:
\begin{equation} 
{\hat{\mathbf{M}}_i} = \sum\nolimits_{j=1}^{N} {\xi_{j \to i}} {\mathbf{M}_i} + \lambda{\mathbf{M}_0}.
\end{equation}
In the DPR module, this study leverages the perceptual and geographical advantages of RSU to assist vehicles in perception fusion. This approach enables the perception model to initially adapt to dynamic environments, achieving a comprehensive perception effect.

\begin{figure}[t]
	\centering
    \includegraphics[width=\linewidth]{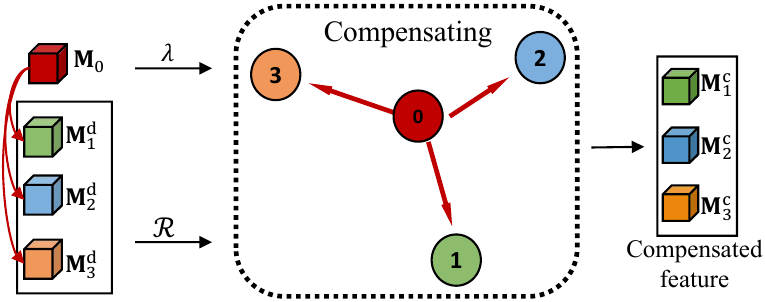}
    \caption{Road-to-Vehicle Perception Compensating. For the feature map $\mathbf{M}_1^{\rm d}$, utilizing RSU-vehicles similarity ratio $\mathcal{R}$ and RSU perception threshold $\lambda$ to determine the extent of RSU perception compensation within $\mathbf{M}_1^{\rm d}$, thus obtaining the compensated feature map $\mathbf{M}_1^{\rm c}$.}
			\label{Fig.4}
 \vspace{-15px}
\end{figure}

\begin{table*}[ht]
  \centering
  \caption{Segmentation comparison of intra-scene. The best options are in bold and ${*}$ indicates the pose-aware version. 
  }
  \vspace{-10px}
  \label{Seg-intra-scene}
  \resizebox*{0.9\linewidth}{!}{
  \begin{tabular}{l|cccccccc|c}
    \toprule
    \textbf{Method}&Unlabeled&Vehicles&Sidewalk&Ground&Road&Buildings&Pedestrian&Vegetation&mIoU(\%)    \\
    \midrule
    Early Fusion&65.96&90.87&94.67&94.52&97.37&94.89&50.45&90.26&84.87   \\
    Late Fusion&48.56&72.17&86.88&85.96&93.48&85.92&18.21&80.07&71.41   \\
    \midrule
    When2com&41.25&65.47&69.62&58.83&83.65&62.36&27.18&62.00&58.79 \\
    When2com*&41.42&63.47&72.19&58.81&81.02&68.55&28.18&74.36&59.75 \\
    Who2com&42.25&66.47&70.62&59.83&84.65&63.36&28.18&63.00&59.80 \\  
    Who2com*&40.02&63.47&72.60&62.81&81.00&60.55&28.20&66.36&60.75 \\
    V2V&60.10&84.92&93.04&91.87&95.98&\textbf{93.10}&33.89&86.85&79.01  \\
    Disco&61.15&84.75&92.82&92.62&96.52&92.95&\textbf{35.49}&\textbf{87.01}&80.41 \\
    \midrule
    AR2VP&\textbf{98.89}&\textbf{85.31}&\textbf{93.37}&\textbf{92.86}&\textbf{96.63}&93.31&33.63&86.38&\textbf{85.05} \\ 
    \bottomrule
\end{tabular}}
\label{Table 1}
\vspace{-15px}
\end{table*}

\noindent
\textbf{Road-to-Vehicle Perception Compensating.}
Due to the continuous changes of scenes, using only the collaborative graph for vehicle perception in dynamic scenarios is insufficient. This paper further leverages the advantages of RSU perceptual stability and extensive coverage to compensate for the updated vehicles perception, thereby enhancing vehicles perception in dynamic scenes.
At this stage, our objective is to utilize the perception features of RSU to compensate for the updated feature maps of vehicles during the decoding process (See Fig.~\ref{Fig.4}).

First, we flat the feature maps of the RSU and vehicles:
\begin{equation}  
{\mathbf{F}^i} = {\rm flatten}({\mathbf{M}}_i^{\rm d}),
\end{equation}
where ${\mathbf{M}}_i^{\rm d}$ is obtained by decoded from ${\hat{\mathbf{M}}_i}$ using a fully-connected layer. 
Then, we calculate the feature similarity ratio (Pearson Correlation Coefficient \cite{per}) ${\mathcal R} = \{r_1,r_2,...,r_i\}$ between the RSU and vehicles, which is to accurately determine which feature map needs to be compensated using RSU perception, avoiding unnecessary computational burdens. 
The formula is as follows:
\begin{equation} 
{r_i} = \frac{\sum\nolimits_{i=1}^{n} {a_i}}{\sqrt {\sum\nolimits_{i=1}^{n} {a_0^2}\cdot{a_i^2} }},
\end{equation}
where
\begin{equation} 
{a_i} = \sum\nolimits_{j=1}^{n} ({\mathbf{F}_j^i} - \sum\nolimits_{i=1}^{k} \frac{\mathbf{M}_i^d}{k}).
\end{equation}
Then, a predefined threshold $\lambda$ is used to determine whether compensation with RSU is required for feature mappings.
We supplement them with RSU perception information, resulting in compensated feature maps ${\mathcal M^{\rm c}}$:
\begin{equation} 
{\mathbf{M}}_i^{\rm c} = (\lambda  - {r_i}){\mathbf{M}}_0^{\rm d} + {\mathbf{M}}_i^{\rm d}.
\end{equation}
Finally, the model is followed by different output heads based on the tasks (such as segmentation (seg.) and detection (det.)) to generate perception results.
Though AR2VP adapts to the changes of intra-scene, but adapting to the changes of inter-scene is still unsolved. 

\begin{figure}[t]
  \centering
  \includegraphics[width=0.8\linewidth]{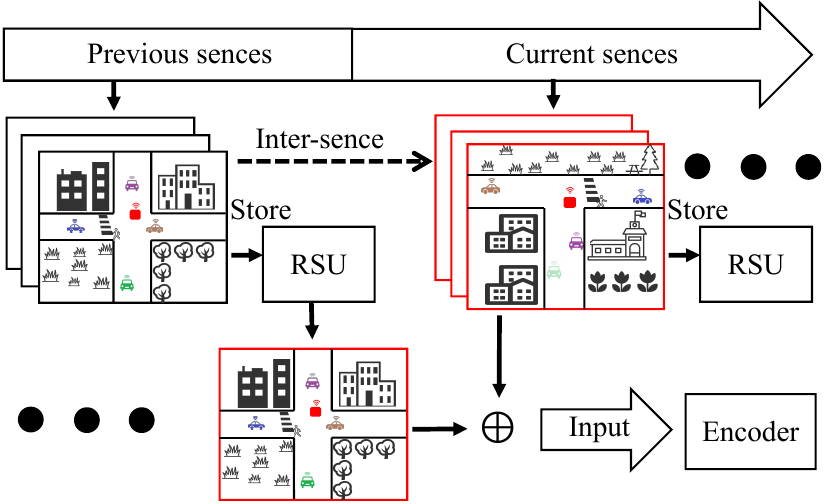}
  \caption{RSU Experience Replay.}
  \label{Fig.5}
 \vspace{-15px}
\end{figure}

\subsection{Overcoming inter-scene changes}

The change of inter-scene is relatively steep, and when the model learns a new scene, it will lead to the forgetting of old scenes knowledge. Inspired by this, this paper introduces \textbf{RSU Experience Replay} (RSU-ER), leveraging RSU storage advantage to retain old scene data and applying continual learning techniques to mitigate forgetting (See Fig.~\ref{Fig.5}).

First, at the learning on new scene data ${\mathcal S_{\rm c}}$, we utilize the storage capability of RSU to store a small set of samples ${\mathcal S_{\rm p}}$:
\begin{equation}
{\mathcal S_{\rm p}} = {\rm Select}(\mu,{\mathcal S_{\rm c}}),
\end{equation}
where, ${\rm Select}(\mu, \cdot )$ represents random selection operation, and $\mu$ is the selected number.
Then, as the model learns new scenes again, we randomly extract a small portion of samples from the ${\mathcal S_{\rm p}}$, concatenate them with the new scene samples for model updating:
\begin{equation}
\theta = {\rm SGD}({\mathcal S_{\rm c}}\cup{\mathcal S_{\rm p}},\theta),
\end{equation}
where, ${\rm SGD}(\cdot)$ is the stochastic gradient descent algorithm, and $\theta$ represents the model parameters.
Lastly, once the model updating is completed, we refresh the ${\mathcal S_{\rm p}}$ once more:
\begin{equation}
{\mathcal S_{\rm p}} = {\rm Select}(\mu,{\mathcal S_{\rm c}})\cup{\mathcal S_{\rm p}}.
\end{equation}
In this stage, the model engages in comprehensive learning of the previous data stored in RSU and the current scene data.
This not only acquires knowledge from new scenes but also revisits knowledge from previous scenes, achieving a synergistic combination of learning and review. Our RSU-ER effectively mitigates the catastrophic forgetting caused by inter-scene changes. 
\subsection{The whole algorithm}
In the process of model learning update to dataset ${\mathcal S}$, we use ${L_{\rm det}}$ loss for the detection \cite{ICCV：det, ICLR：det} task to update learning:
\begin{equation} 
L_{\rm det} = \sum\nolimits_{i=1}^{n} {\frac{\eta ({Y_i-Y_i'})^2}{\sigma^2}},
\end{equation}
where, $\eta$ typically takes a value of 0.5, in the segmentation \cite{ICML1：seg,ICML2：seg} task, we use ${L_{\rm seg}}$ loss for update learning:

\begin{equation} 
L_{\rm seg} =  - \sum\nolimits_{i=1}^{n} (Y_i\cdot\log (Y_i')) ,
\end{equation}
where ${Y}$ and ${Y'}$ represent the label and prediction in scene ${\mathcal S}$, ${\sigma}$ is a hyperparameter.

The overall update loss ${L}$ of the model is as follows:
\begin{equation}
L = L_{\rm det/seg}^{\rm previous} + L_{\rm det/seg}^{\rm current} ,
\end{equation}
where $L_{\rm det/seg}^{\rm previous}$ represents the loss from replayed previous scene data used for the current task $(\rm det./seg.)$, and $L_{\rm det/seg}^{\rm previous}$ represents the loss from current scene data used for the current task.

In the whole AR2VP research (See Fig.~\ref{Fig.2}), we design DPR module (See Fig.~\ref{Fig.3}), merging geographical and feature data from RSU and vehicles to create an adaptable collaborative graph for dynamic scenarios.
This effectively integrates perception information from different vehicles, enabling a more comprehensive grasp of dynamic elements within the scene. 
Subsequently, inspired by residual techniques, we propose the R2VPC module (See Fig.~\ref{Fig.4}). By leveraging RSU perceptual advantages, this module compensates post-collaborative vehicle perception, filling in intra-scene dynamic elements overlooked by the vehicles, further enhancing overall adaptability to dynamic settings. Lastly, to extend adaptability beyond intra-scene changes, we introduce RSU-ER (See Fig.~\ref{Fig.5}), combining RSU storage capacity and experience replay techniques. This empowers AR2VP to cope with inter-scene changes, ensuring robust and reliable vehicle perception. 
Note that AR2VP also considers \emph{to save the communication bandwidth}, where RSU and vehicles could compress their feature map prior to transmission. 
Optionally, in our study, we make use a 1 × 1 convolutional autoencoder \cite{compress} to compress and decompress the feature maps along the channel dimension. The autoencoder is trained together with the whole system.


\begin{figure*}[t]
  \centering
  \includegraphics[width=\linewidth]{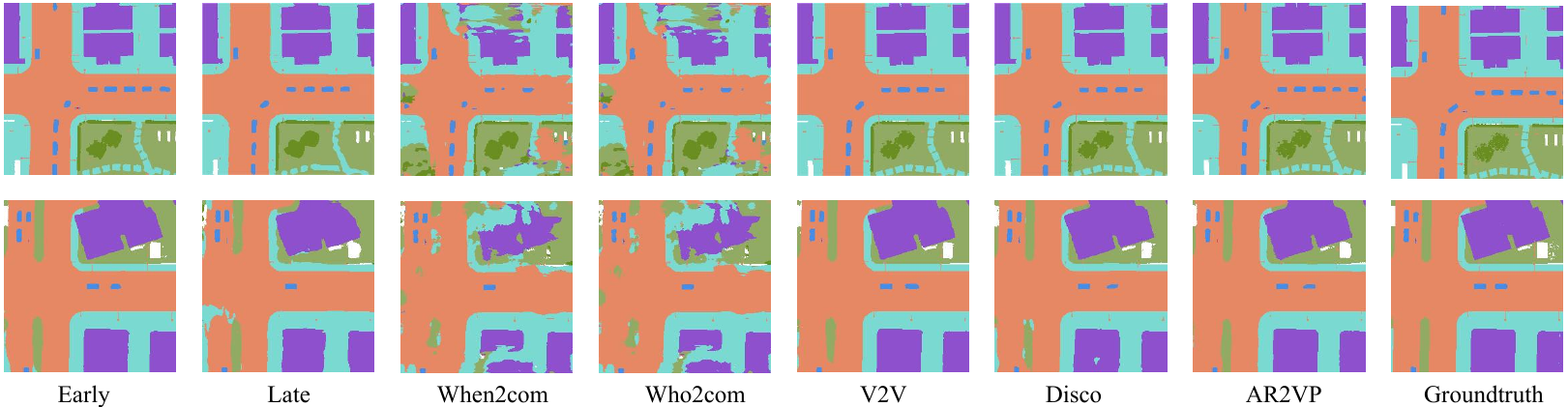}
  \caption{Visualizations of collaborative BEV semantic segmentation.}
  \label{Fig.6}
\end{figure*}

\begin{figure*}[t]
  \centering
  \includegraphics[width=\linewidth]{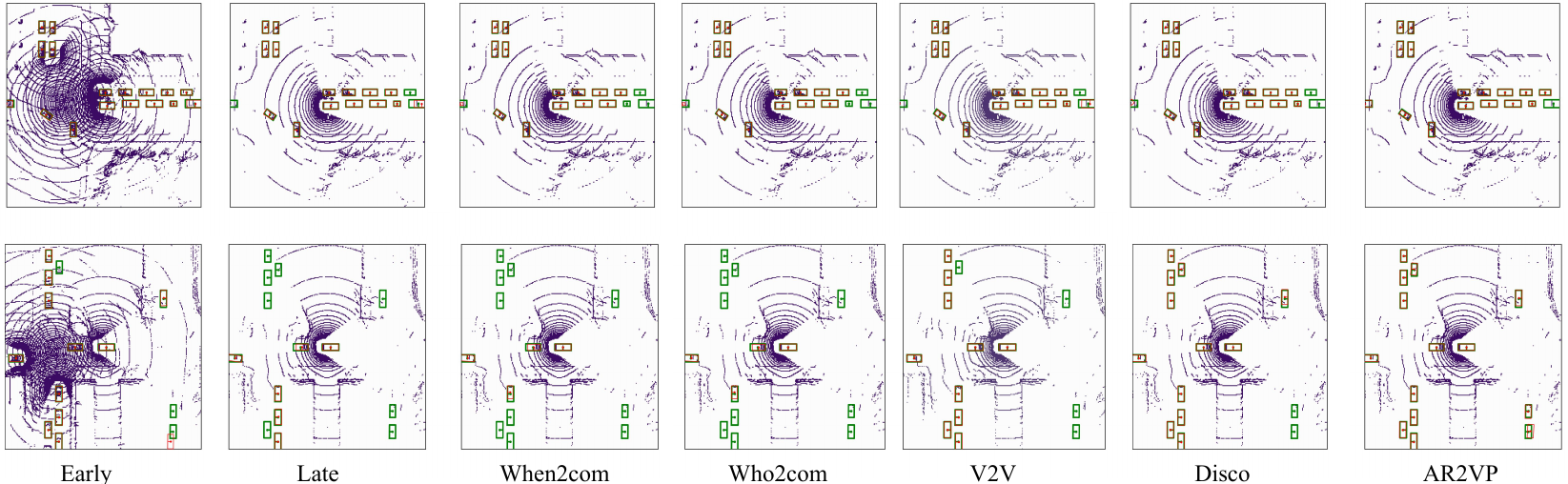}
  \caption{Visualizations of BEV detection on V2X-Sim. Red and green boxes are the predictions and ground-truths respectively.}
  \label{Fig.7}
  \vspace{-15px}
\end{figure*}

\section{Experiment}

\subsection{Data preparation and evaluation metric}

In this study, we employ the V2X-sim dataset to evaluate the V2X perception task. The V2X-sim dataset emulates multi-agent scenarios, wherein each scenario encompasses a 20-second traffic flow across multiple intersections. Laser radar recordings are captured at intervals of 0.2 seconds, yielding a total of 100 frames per scenario. This dataset comprises 100 distinct scenes, with each frame housing multiple samples. The training set comprises 23,500 samples, while the test set contains 3,100 samples.
To establish a fixed large scenario, we selected 30 scenes, which collectively contribute 3,000 frames. Among these, the training set comprises 2,700 frames, while the test set consists of 300 frames. 
Moreover, to implement cross-scene experiments, we also train V2X model sequentially on three major scenes in chronological order.

\begin{table}[t]
\centering
  \caption{Detection comparison of intra-scene.}
  \vspace{-10px}
  \label{tab:freq}
  \resizebox*{0.6\linewidth}{!}{
  \begin{tabular}{l|cc}
   \toprule
   \multirow{2}*{\textbf{Method}}&\multicolumn{2}{c}{mAP(\%)} \\
   &AP@0.5&AP@0.7 \\
   \midrule
   Early Fusion&96.63&96.05 \\
   Late Fusion&85.62&83.84 \\
   \midrule
   When2com&81.35&80.02 \\
   When2com*&81.86&80.69 \\
   Who2com&81.32&79.98 \\
   Who2com*&81.69&80.66 \\
   V2V&91.89&89.90 \\
   Disco&92.01&90.41 \\
   \midrule
   AR2VP&\textbf{94.50}&\textbf{92.77} \\
   \bottomrule
\end{tabular}}
\label{Det-intra-scene}
\vspace{-15px}
\end{table}

In this paper, we evaluate our method on two V2X perception tasks, including scene segmentation and vehicle objection. 
We employ the generic BEV detection evaluation metric: \textit{Average Precision} (AP) at \textit{Intersection-over-Union} (IoU) threshold of 0.5 and 0.7. We evaluate the segmentation performance using mean IoU (mIoU). We evaluate the extent of forgetting across inter-scene changes using Forget. 

\subsection{Quantitative evaluation}

\noindent
\textbf{Compared methods.} 
We first compare with the early collaboration method \cite{64} and the late collaboration method, which are always seems as the upper bound and lower in traditional V2X perception tasks.
Furthermore, four intermediate collaboration methods are used, inculuding When2com \cite{18}, Who2com \cite{19}, V2V \cite{v2vnet} and Disco \cite{62}. Since the original Who2com and When2com do not consider pose information, we consider both pose-aware and pose-agnostic versions (with $*$) to achieve fair comparisons. All the methods use the same segmentation and detection backbones and conduct collaboration at the same intermediate feature layer.

\noindent
\textbf{Comparisons under intra-sence changes.} 
Tables \ref{Seg-intra-scene} and \ref{Det-intra-scene} show the comparisons in terms of mIoU (seg.) and AP@0.5/0.7 (det.). 
Comparing to the pose-aware When2com, AR2VP improves by 57.57\% in segmentation of unlabeled data and 26.26\% in mIoU. 
Comparing to Disco, AR2VP improves by 4.64\%. 
Comparing to the pose-aware When2com, AR2VP improves by 13.15\% in AP@0.5 and 12.75\% in AP@0.7. 
Comparing to Disco, AR2VP improves by 3.4\% in AP@0.5 and 3.2\% in AP@0.7. 
The qualitative results are shown in Fig. \ref{Fig.6} (seg.) and Fig.~\ref{Fig.7} (det.). 
We observed that AR2VP demonstrates superior entity perception outcomes, achieving the highest overall perception performance. This analysis underscores that current V2X technologies rarely rely on RSUs to expand perception horizons. In contrast, AR2VP harnesses the latent strengths of RSUs to address intra-scene changes, which enhances the vehicle's ability to adapt to dynamic scenes, consequently elevating the overall perception capabilities. However, AR2VP does exhibit a performance drawback in pedestrian detection, implying a particular challenge in detecting small targets.

\begin{figure}[t]
  \centering
  \includegraphics[width=0.8\linewidth]{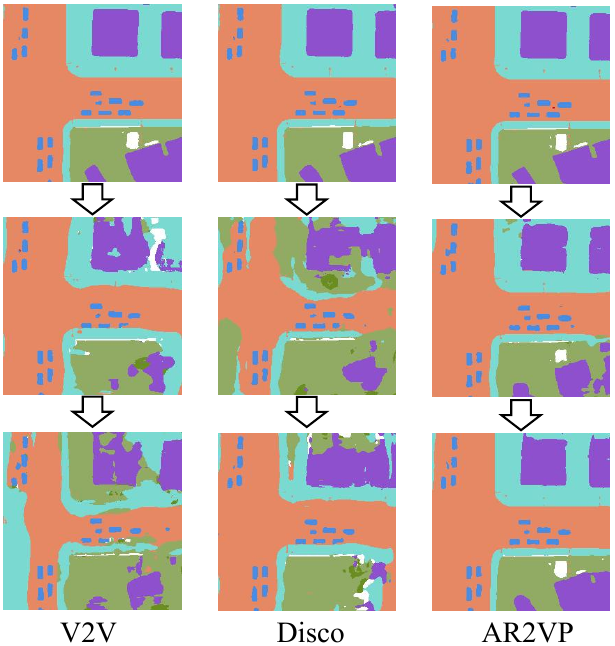}
  \caption{Visualizations of collaborative BEV semantic segmentation on  inhibition of forgetting. 
  }
  \label{Fig.8}
\end{figure}

\noindent
\textbf{Comparisons under inter-scene change.} Table \ref{inter-scene} shows the comparison on inhibition of forgetting cross different scenes. Comparing to the pose-aware When2com, AR2VP improves by 30.78\% in mIoU and reduce forgetting rate by 19.28\%. Comparing to Disco, AR2VP improves by 20.42\% in mIoU and reduce forgetting rate by 23.42\%; Comparing to the pose-aware When2com, AR2VP improves by 28.12\% in AP@0.5 and 27.00\% in AP@0.7, and reduce forgetting rate by 26.07\% in AP@0.5 and 25.60\% in AP@0.7. Comparing to Disco, AR2VP improves by 6.05\% in AP@0.5 and 6.44\% in AP@0.7, and reduce forgetting rate by 8.23\% in AP@0.5 and 9.94\% in AP@0.7. 
AR2VP presents itself as a frontrunner in terms of overall perception performance. Upon analysis, it's evident that traditional V2X technologies disregard the influence of inter-scene changes on perception. In contrast, AR2VP optimally exploits the storage capacity of RSU and integrates continuous learning principles to effectively address inter-scene changes. This strategic approach empowers vehicles to assimilate new scenes while minimizing the extent of memory loss from prior scenes. This capability shows a strong adaptability to inter-scene changes in perception, thereby enhancing the global robustness of perception.
Although RSU-ER can be applied to other models to mitigate forgetting, AR2VP notably demonstrates the most favorable suitability. Fig.~\ref{Fig.8} portrays the learning of new scene data based on the old model, revealing the degree of memory forgetting from previous scenes. The observation is clear: V2V and Disco struggle to accommodate inter-scene changes, leading to significant memory loss from previous scenes. In contrast, AR2VP adeptly navigates inter-scene changes, exhibiting a higher retention of memories from prior scenes. This analysis underscores AR2VP's capacity for the lowest forgetting rate and the most proficient performance in addressing inter-scene changes.


\begin{figure}[t]
  \centering
  \includegraphics[width=\linewidth]{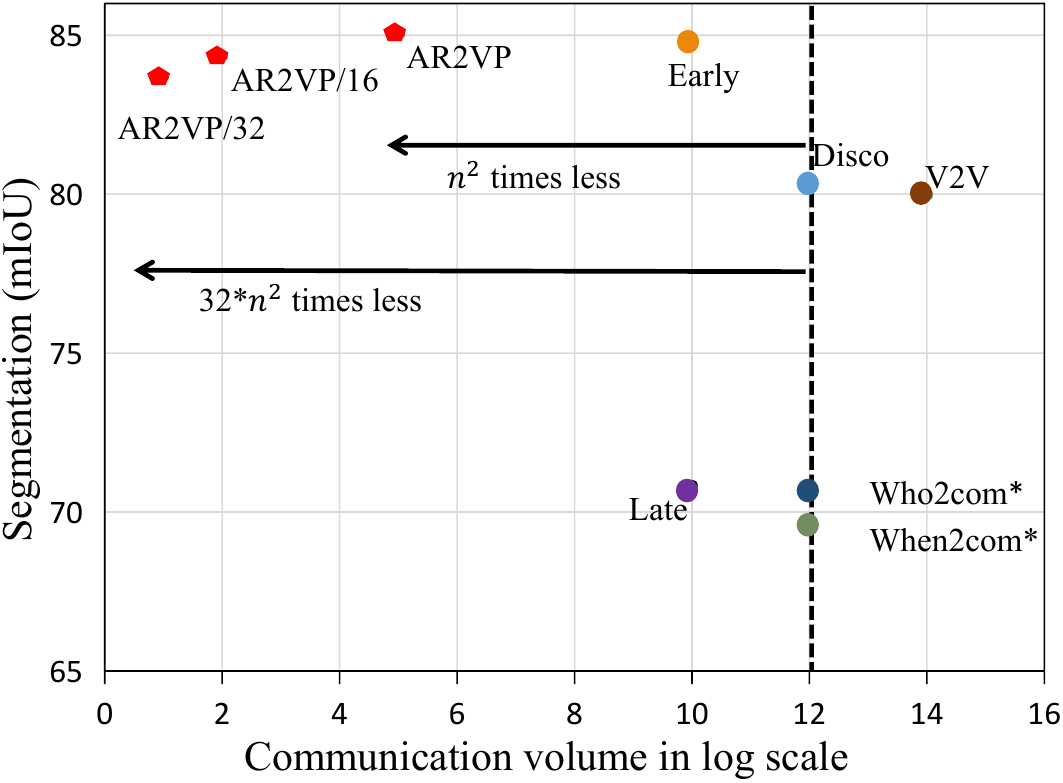}
  \caption{Performance-bandwidth trade-off. 
  }
  \label{Fig.9}
\vspace{-15px}
\end{figure}

\noindent
\textbf{Performance-bandwidth trade-off analysis.}
In Fig.~\ref{Fig.9}, we compare the proposed AR2VP with the baseline methods in terms of the trade-off between segmentation performance and communication bandwidth. The dashed line represents the baseline based on when2com.
To show the better trade-off of the proposed AR2VP, we employ an autoencoder to compress features and reduce the communication bandwidth used for feature transmission ($/n$ means compress $n$ times).
We have the following observations:
1) Comparing to AR2VP, AR2VP/32 degrades by 1.31\% in segmentation, still outperforming Disco in terms of mIoU. 
This means the compressing features in AR2VP does not significantly compromise perception performance.
2) Storing the model in the RSU further reduces ${n^2}$ communication bandwidth, where ${n}$ is the number of vehicles participating in the collaboration.

\begin{table}[t]
  \centering
  \caption{Perception comparisons on inter-scene changes. 
  }
  \vspace{-10px}
  \label{tab:freq}
  \resizebox*{0.97\linewidth}{!}{
  \begin{tabular}{l|cc|cc|cc}
   \toprule
   \multirow{2}*{\textbf{Method}}&\multicolumn{2}{|c}{Det.(AP@0.5(\%))}&\multicolumn{2}{|c}{Det.(AP@0.7(\%))}&\multicolumn{2}{|c}{Seg.(\%)}\\
   &mAP↑ &Forget↓ &mAP↑ &Forget↓ &mIoU↑ &Forget↓ \\
   \midrule
   Early Fusion&85.15&14.96&81.03&19.19&52.71&42.62 \\
   Late Fusion&62.77&27.61&57.86&31.96&4014&31.91 \\
   \midrule
   When2com&61.40&32.23&57.90&34.44&40.73&30.76 \\
   When2com*&61.64&31.38&57.98&33.01&45.60&28.65 \\
   Who2com&61.04&32.89&57.66&34.63&41.63&31.56 \\
   Who2com*&61.47&32.01&58.11&33.79&44.71&29.56 \\
   V2V&82.54&15.65&76.98&20.63&48.50&46.08 \\
   Disco&83.47&14.39&78.46&18.78&48.09&43.31 \\
   \midrule
   V2V (RSU-ER)&87.63&7.38&83.48&9.46&63.06&18.09 \\
   Disco (RSU-ER)&87.65&7.61&83.65&9.53&64.74&19.12 \\
    \midrule
   AR2VP (RSU-ER)&\textbf{89.52}&\textbf{6.16}&\textbf{84.90}&\textbf{8.84}&\textbf{71.51}&\textbf{11.48} \\
   \bottomrule
   \end{tabular}
}
\label{inter-scene}
\end{table}

\begin{table}[t]
  \centering
  \caption{Ablation studies.}
  \vspace{-10px}
  \label{tab:freq}
  \resizebox*{0.9\linewidth}{!}{
  \begin{tabular}{c|cc|cc|c}
   \toprule
   \multirow{2}*{\textbf{RSU}} &\multirow{2}*{\textbf{graph}} & \multirow{2}*{\textbf{compensator}}&\multicolumn{2}{|c|}{Det.(\%)}&Seg.(\%)\\
   &&&Ap@0.5&Ap@0.7&mIoU \\
   \midrule
   \XSolidBrush&\XSolidBrush&\Checkmark&66.99&65.47&55.01\\
   \XSolidBrush&\Checkmark&\XSolidBrush &89.88&87.95&75.65 \\
   \XSolidBrush&\Checkmark&\Checkmark &90.65&88.46&73.06 \\
   \midrule
   \Checkmark&\XSolidBrush&\Checkmark &68.56&66.32&56.47\\
   \Checkmark&\Checkmark&\XSolidBrush &93.80&91.71&84.04 \\
   \Checkmark&\Checkmark&\Checkmark& \textbf{94.50}&\textbf{92.77}&\textbf{85.05} \\
   \bottomrule
\end{tabular}
}
\label{Ablation}
\vspace{-15px}
\end{table}

 \noindent
\textbf{Ablation study.}
We conduct ablation studies to analyze the perceptual performance of graph, and the communication in the presence and absence of RSU. 
The results are shown in Table \ref{Ablation}.
First, we find that the participation of RSU in the collaborative process provide additional perception coverage to enhance vehicle perception performance.
Second, the collaborative graph effectively integrates all perception information, enabling vehicles to comprehensively perceive entities within the scene. 
Third, in scenarios where RSU is present, the compensator utilize the stable perception information from RSU to efficiently compensate for vehicle perception. 
Moreover, the compensator benefits from the presence of RSU, showing a positive effect. In the absence of RSU, using vehicles to compensate for other vehicles' perception would lead to negative consequences.

\section{Conclusion}
In this paper, we proposed a vehicle-road cooperative perception model, named AR2VP, which is capable of adapting to dynamic environments. It mainly consists of a DPR module, a R2VPC module and RSU-ER method. 
The DPR module efficiently integrates vehicle perceptions to comprehensively capture dynamic factors within the scene, enhancing the perception capabilities of the collaborative perception model.
The R2VPC module is geared towards effectively retaining the optimal RSU perception information, especially in the face of intra-scene changes.
The RSU-ER method integrates within the RSU's storage capacity, facilitates the retention of a small volume of historical scene data. This approach ensures that the cooperative model maintains a certain level of robustness when confronted with inter-scene changes.
Comprehensive experiments demonstrate that AR2VP achieves adaptability to dynamic environments and an appealing performance-bandwidth trade-off through a more direct design principle. 
In the future, based on our experimental findings, we intend to enhance the AR2VP's capability in recognizing small objects. This will involve further refinement of the model to ensure more accurate and effective identification of small entities.

\bibliography{./aaai24}

\begin{thebibliography}{40}
\providecommand{\natexlab}[1]{#1}

\bibitem[{A.Demba and D.P.F.Möller(2018)}]{43(v2v)}
A.Demba; and D.P.F.Möller. 2018.
\newblock {Vehicle-to-Vehicle Communication Technology}.
\newblock In \emph{EIT}.

\bibitem[{A.Geiger(2012)}]{2}
A.Geiger, R., P.Lenz. 2012.
\newblock {Are we ready for autonomous driving? the KITTI vision benchmark suite}.
\newblock In \emph{CVPR}.

\bibitem[{C.Hao(2023)}]{58.2}
C.Hao, L., L.LinYan. 2023.
\newblock {Multi-semantic hypergraph neural network for effective few-shot learning}.
\newblock \emph{PR}.

\bibitem[{C.Qi and F.Song(2019)}]{64}
C.Qi, Y., T.Sihai; and F.Song. 2019.
\newblock {Cooper: Cooperative perception for connected autonomous vehicles based on 3d point clouds}.
\newblock In \emph{ICDCS}.

\bibitem[{C.Shao and et~al(2022)}]{ACL2：forget}
C.Shao, Y.; and et~al. 2022.
\newblock {Overcoming Catastrophic Forgetting beyond Continual Learning: Balanced Training for Neural Machine Translation}.
\newblock In \emph{ACL}.

\bibitem[{C.Yiming(2023)}]{ICML1：seg}
C.Yiming, Y., Y.Linjie. 2023.
\newblock {Learning Dynamic Query Combinations for Transformer-based Object Detection and Segmentation}.
\newblock In \emph{ICML}.

\bibitem[{D.Jiahua and S.Gan(2023)}]{incremental1}
D.Jiahua, C., L.Wenqi; and S.Gan. 2023.
\newblock {Heterogeneous Forgetting Compensation for Class-Incremental Learning}.
\newblock In \emph{ICCV}.

\bibitem[{D.Kaile and et~al(2023)}]{58.3}
D.Kaile, L.; and et~al. 2023.
\newblock {Multi-Label Continual Learning using Augmented Graph Convolutional Network}.
\newblock \emph{TMM}.

\bibitem[{D.Wenlu and et~al(2023)}]{AAAI3：res}
D.Wenlu, G., Y.Junyi; and et~al. 2023.
\newblock {SafeLight: A Reinforcement Learning Method toward Collision-free Traffic Signal Control}.
\newblock In \emph{AAAI}.

\bibitem[{E.Verwimp and et~al(2022)}]{ICCV：det}
E.Verwimp, S., K.Yang; and et~al. 2022.
\newblock {CLAD: A realistic Continual Learning benchmark for Autonomous Driving}.
\newblock In \emph{ICLR}.

\bibitem[{G.Winata and et~al(2023)}]{ACL1：forget}
G.Winata, K., L.Xie; and et~al. 2023.
\newblock {Overcoming Catastrophic Forgetting in Massively Multilingual Continual Learning}.
\newblock In \emph{ACL}.

\bibitem[{Ha.Wang and Y.Cai(2022)}]{45(v2i)}
Ha.Wang, X.; and Y.Cai. 2022.
\newblock {V2I-CARLA: A Novel Dataset and a Method for Vehicle Reidentification-Based V2I Environment}.
\newblock \emph{IEEE TIM}.

\bibitem[{H.Kai and G.Yutao(2023)}]{incremental}
H.Kai, X., W.Feigege; and G.Yutao. 2023.
\newblock {Prototypical Kernel Learning and Open-set Foreground Perception for Generalized Few-shot Semantic Segmentation}.
\newblock In \emph{ICCV}.

\bibitem[{H.TianZhang and et~al(2023)}]{RSU2}
H.TianZhang, N., Adel; and et~al. 2023.
\newblock {An Intent-based Framework for Vehicular Edge Computing}.
\newblock In \emph{PerCom}.

\bibitem[{H.Xie and et~al(2023)}]{ICML2：seg}
H.Xie, K., J.Zhu; and et~al. 2023.
\newblock {A Critical View of Vision-Based Long-Term Dynamics Prediction Under Environment Misalignment}.
\newblock In \emph{ICML}.

\bibitem[{L.Fan and et~al(2023)}]{58}
L.Fan, S.; and et~al. 2023.
\newblock {Measuring Asymmetric Gradient Discrepancy in Parallel Continual Learning }.
\newblock In \emph{ICCV}.

\bibitem[{L.Fan and et~al(2021)}]{59}
L.Fan, W.; and et~al. 2021.
\newblock {Multi-Domain Multi-Task Rehearsal for Lifelong Learning}.
\newblock In \emph{AAAI}.

\bibitem[{L.Hao and C.Xiong(2019)}]{61}
L.Hao, A.; and C.Xiong. 2019.
\newblock {Competitive Experience Replay}.
\newblock In \emph{ICLR}.

\bibitem[{L.Sebastian and P.A.Hooper(2023)}]{ICLR：det}
L.Sebastian; and P.A.Hooper. 2023.
\newblock {In-situ Anomaly Detection in Additive Manufacturing with Graph Neural Networks}.
\newblock In \emph{ICLR}.

\bibitem[{L.Yiming and et~al(2021)}]{62}
L.Yiming, W., R.Shunli; and et~al. 2021.
\newblock {Learning Distilled Collaboration Graph for Multi-Agent Perception}.
\newblock In \emph{NIPS}.

\bibitem[{M.Hasan and H.Lu(2018)}]{9}
M.Hasan, T., S.Mohan; and H.Lu. 2018.
\newblock {Securing vehicle-to-everything (V2X) communication platforms}.
\newblock \emph{IEEE Trans. Intell. Veh.}

\bibitem[{M.Jonathan and S.Jürgen(2011)}]{compress}
M.Jonathan, C., M.Ueli; and S.Jürgen. 2011.
\newblock {FrozenRecon: Pose-free 3D Scene Reconstruction with Frozen Depth Models}.
\newblock In \emph{ICANN}.

\bibitem[{M.Muhammad and G.A.Safdar(2018)}]{8}
M.Muhammad; and G.A.Safdar. 2018.
\newblock {Survey on existing authentication issues for cellular-assisted V2X communication}.
\newblock \emph{Vehicle Communication}.

\bibitem[{M.van and S.Tolias(2020)}]{60}
M.van, T.; and S.Tolias. 2020.
\newblock {Brain-inspired replay for continual learning with artificial neural networks}.
\newblock \emph{Nature Communications}.

\bibitem[{R.Xu and J.Ma(2022)}]{20}
R.Xu, X., H.Xiang; and J.Ma. 2022.
\newblock {OPV2V: An open benchmark dataset and fusion pipeline for perception with vehicle-to-vehicle communication}.
\newblock In \emph{ICRA}.

\bibitem[{S.Anuradha and et~al(2023)}]{per}
S.Anuradha, Y.; and et~al. 2023.
\newblock {Linking Alternative Fuel Vehicles Adoption with Socioeconomic Status and Air Quality Index}.
\newblock In \emph{AAAI}.

\bibitem[{S.Peiyuan(2023)}]{RSU1}
S.Peiyuan, Z., Q.Liangxin. 2023.
\newblock {A Hybrid Framework of Reinforcement Learning and Convex Optimization for UAV-Based Autonomous Metaverse Data Collection}.
\newblock \emph{IEEE Network magazine}.

\bibitem[{S.Qing and et~al(2023)}]{58.1}
S.Qing, L.; and et~al. 2023.
\newblock {Exploring Example Influence in Continual Learning}.
\newblock In \emph{NeurIPS}.

\bibitem[{T.Ren(2023)}]{AAAI1：conti}
T.Ren, Z. 2023.
\newblock {Integrating Curricula with Replays: Its Effects on Continual Learning}.
\newblock In \emph{AAAI}.

\bibitem[{T.Wang and et~al(2020)}]{v2vnet}
T.Wang, L., S.Manivasagam; and et~al. 2020.
\newblock {V2vnet: Vehicle-to-vehicle communication for joint perception and prediction}.
\newblock In \emph{ECCV}.

\bibitem[{V.Nicholas and et~al(2020)}]{65}
V.Nicholas, R.; and et~al. 2020.
\newblock {Learning to communicate and correct pose errors}.
\newblock In \emph{CoRL}.

\bibitem[{W.Chenglong and et~al(2023)}]{freez}
W.Chenglong, Z., Y.Jiangyan; and et~al. 2023.
\newblock {Low-rank Adaptation Method for Wav2vec2-based Fake Audio Detection}.
\newblock In \emph{IJCAI}.

\bibitem[{X.Guangkai and et~al(2023)}]{freez1}
X.Guangkai, G., Y.Wei; and et~al. 2023.
\newblock {FrozenRecon: Pose-free 3D Scene Reconstruction with Frozen Depth Models}.
\newblock In \emph{ICCV}.

\bibitem[{Y.-C.Liu and N.Glaser(2020{\natexlab{a}})}]{18}
Y.-C.Liu, J.; and N.Glaser. 2020{\natexlab{a}}.
\newblock {When2com: Multi-agent perception via communication graph grouping}.
\newblock In \emph{CVPR}.

\bibitem[{Y.-C.Liu and N.Glaser(2020{\natexlab{b}})}]{19}
Y.-C.Liu, J.; and N.Glaser. 2020{\natexlab{b}}.
\newblock {Who2com: Collaborative perception via learnable handshake communication}.
\newblock In \emph{ICRA}.

\bibitem[{Y.~Li and Wang(2022)}]{63}
Y.~Li, Z.~A., Dekun~Ma; and Wang, Z. 2022.
\newblock {V2X-Sim: Multi-Agent Collaborative Perception Dataset and Benchmark for Autonomous Driving}.
\newblock \emph{RAL}.

\bibitem[{Y.Li and et~al(2021)}]{13}
Y.Li, P., S.Ren; and et~al. 2021.
\newblock {Learning distilled collaboration graph for multi-agent perception}.
\newblock In \emph{NIPS}.

\bibitem[{Y.Yuan and M.Sester(2021)}]{14}
Y.Yuan; and M.Sester. 2021.
\newblock Comap: A synthetic dataset for collective multiagent perception of autonomous driving.
\newblock In \emph{NIPS}.

\bibitem[{Z.Jiarui and et~al(2023)}]{AAAI4：dri}
Z.Jiarui, K., I.Filip; and et~al. 2023.
\newblock {Utilizing Background Knowledge for Robust Reasoning over Traffic Situations}.
\newblock In \emph{AAAI}.

\bibitem[{Z.Lei and et~al(2023)}]{AAAI2：res}
Z.Lei, H., Y.Xiaodong; and et~al. 2023.
\newblock {DRGCN: Dynamic Evolving Initial Residual for Deep Graph Convolutional Networks}.
\newblock In \emph{AAAI}.

\end{thebibliography}



\end{document}